\documentclass[lettersize,journal]{IEEEtran}
\usepackage{amsmath,amsfonts}
\usepackage{amssymb}
\usepackage{algorithmic}
\usepackage{algorithm}
\usepackage{array}
\usepackage[caption=false,font=normalsize,labelfont=sf,textfont=sf]{subfig}
\usepackage{textcomp}
\usepackage{stfloats}
\usepackage{url}
\usepackage{verbatim}
\usepackage{graphicx}
\usepackage{placeins}
\usepackage{cite}
\usepackage{amsmath}
\usepackage{mathtools}
\usepackage{booktabs}
\usepackage{multirow}
\usepackage{xr}
\makeatletter
\newcommand*{\addFileDependency}[1]{%
  \typeout{(#1)}%
  \@addtofilelist{#1}%
  \IfFileExists{#1}{}{\typeout{No file #1.}}%
}
\makeatother

\addFileDependency{main.aux}

\begin{document}

\title{SynCast: Synergizing Contradictions in Precipitation Nowcasting via Diffusion Sequential Preference Optimization}



\author{Kaiyi Xu, Junchao Gong, Wenlong Zhang, Ben Fei, Lei Bai, Wanli Ouyang
\thanks{This work was supported by Shanghai Artificial Intelligence Laboratory the JC STEM Lab of AI for Science and Engineering, funded by The Hong Kong Jockey Club Charities Trust,  the MTR Research Funding (MRF) Scheme (CHU-24003), the Research Grants Council of Hong Kong (Project No. CUHK14213224).~(\textit{Corresponding author: Ben Fei}).}
\thanks{Kaiyi Xu is with the School of Information Science and Technology, University of Science and Technology of China (USTC), Anhui, China. (e-mail: xuky@mail.ustc.edu.cn), and also with Shanghai Artificial Intelligence Laboratory, Shanghai 200232, China. This work was done during her internship at Shanghai Artificial Intelligence Laboratory.}
\thanks{Junchao Gong is with the School of Electronic Information and Electrical Engineering, Shanghai Jiao Tong University, Shanghai 200240, China, and also with Shanghai Artificial Intelligence Laboratory, Shanghai 200232, China (e-mail: gjchimself@sjtu.edu.cn).}
\thanks{Wenlong Zhang and Lei Bai are with the Shanghai Artificial Intelligence Laboratory, Shanghai 200232, China (email: zhangwenlong@pjlab.org.cn; bailei@pjlab.org.cn).}

\thanks{Ben Fei and Wanli Ouyang are with the Chinese University of Hong Kong (email: benfei@cuhk.edu.hk; wlouyang@ie.cuhk.edu.hk).}
}

\markboth{Journal of \LaTeX\ Class Files,~Vol.~14, No.~8, August~2021}%
{Shell \MakeLowercase{\textit{et al.}}: A Sample Article Using IEEEtran.cls for IEEE Journals}


\maketitle

\begin{abstract}
Precipitation nowcasting based on radar echoes plays a crucial role in monitoring extreme weather and supporting disaster prevention.
Although deep learning approaches have achieved significant progress, they still face notable limitations. 
For example, deterministic models tend to produce over-smoothed predictions, which struggle to capture extreme events and fine-scale precipitation patterns. 
Probabilistic generative models, due to their inherent randomness, often show fluctuating performance across different metrics and rarely achieve consistently optimal results.
Furthermore, precipitation nowcasting is typically evaluated using multiple metrics, some of which are inherently conflicting. 
For instance, there is often a trade-off between the Critical Success Index (CSI) and the False Alarm Ratio (FAR), making it challenging for existing models to deliver forecasts that perform well on both metrics simultaneously.
To address these challenges, we introduce preference optimization into precipitation nowcasting for the first time, motivated by the success of reinforcement learning from human feedback in large language models.
Specifically, we propose SynCast, a method that employs the two-stage post-training framework of Diffusion Sequential Preference Optimization (Diffusion-SPO), to progressively align conflicting metrics and consistently achieve superior performance.
In the first stage, the framework focuses on reducing FAR, training the model to effectively suppress false alarms. 
Building on this foundation, the second stage further optimizes CSI with constraints that preserve FAR alignment, thereby achieving synergistic improvements across these conflicting metrics.
Experiments on three radar precipitation datasets demonstrate that SynCast reduces FAR while improving CSI, and achieves performance comparable to state-of-the-art methods. 
Furthermore, we verify that the post-training framework of Diffusion-SPO is compatible with multiple diffusion models for precipitation nowcasting, demonstrating its generalizability.
\end{abstract}

\begin{figure}[ht]
    \centering
    \includegraphics[width=1\linewidth]{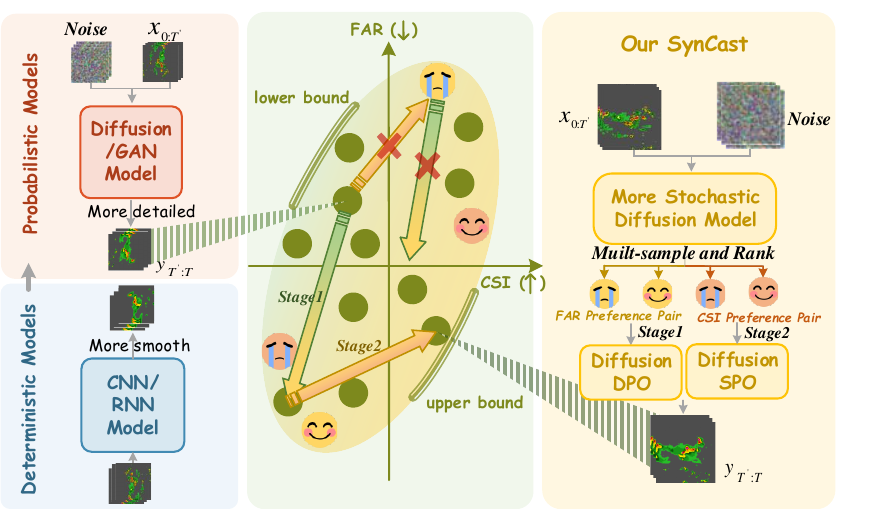}
    \caption{Different precipitation nowcasting methods. Compared with existing deterministic and probabilistic approaches for precipitation nowcasting, SynCast leverages a probabilistic generative model combined with \textbf{reinforcement learning fine-tuning} to preserve fine-grained details while approaching the upper performance bounds of the conflicting metrics FAR and CSI.}
    \label{fig:teaser}
\end{figure} 
\section{Introduction}
\IEEEPARstart{P}{recipitation} nowcasting based on radar observations is a fundamental task in weather prediction~\cite{ravuri2021skilful, lin2005precipitation}. 
Its accuracy directly affects early warnings of extreme weather events and the effectiveness of public safety protection. 
With the increasing frequency and intensity of extreme weather events in recent years, this task has received growing attention from worldwide researchers~\cite{yao2022emergence, howe2021extreme}.

Previous approaches largely relied on deterministic models~\cite{gao2022earthformer, wang2022predrnn, gao2022simvp}, which optimize mean error-based objectives (e.g., Root Mean Squared Error or Mean Squared Error) and consequently tend to produce overly smoothed predictions.

To address this limitation, recent work has turned to probabilistic generative models, such as Diffusion Models and GANs~\cite{yu2024diffcast, gong2024cascast, zhang2023skilful}. 
Rather than producing a single deterministic prediction, these models aim to model the conditional distribution of future precipitation.
By sampling in the latent or noise space, they introduce randomness that enables the representation of inherent uncertainty in future weather systems.
However, this randomness also brings fluctuations in performance when evaluated on various metrics, as different random seeds may partially alter the spatial structure and intensity of predictions.
Consequently, the performance of probabilistic models is naturally bounded. Existing approaches usually vary within this range and rarely achieve the upper bound consistently~\cite{yu2024diffcast, gao2023prediff, rombach2022high} as shown in Fig.~\ref{fig:teaser}. 
To mitigate the instability caused by random seeds, ensemble forecasting has been employed~\cite{parker2013ensemble, lin2024enhancing}. 
Although combining multiple predictions could capture uncertainty, this averaging process also introduces the smoothing problem observed in deterministic models.

A similar challenge also arises in Large Language Models (LLMs)~\cite{li2024dissecting, zhou2024fairer}: the same prompt can generate responses of varying quality, and only a subset satisfies human preferences.
Therefore, preference optimization~\cite{ouyang2022training, rafailov2023direct, meng2024simpo} addresses this by refining the model with paired preference data, making the response that humans prefer more likely under sampling. 
This strategy can be naturally extended to probabilistic precipitation nowcasting due to its stochastic nature.
Specifically, paired preference data can be constructed based on domain-specific evaluation metrics, such as the CSI and FAR, and a probabilistic model is then fine-tuned within a Diffusion Direct Preference Optimization (DPO) framework~\cite{rafailov2023direct} to favor ``win'' samples while suppressing ``lose'' samples.
As a result, the model can approach the upper bound of probabilistic performance under a single evaluation metric.

\begin{figure}[ht]
    \centering
    \includegraphics[width=1\linewidth]{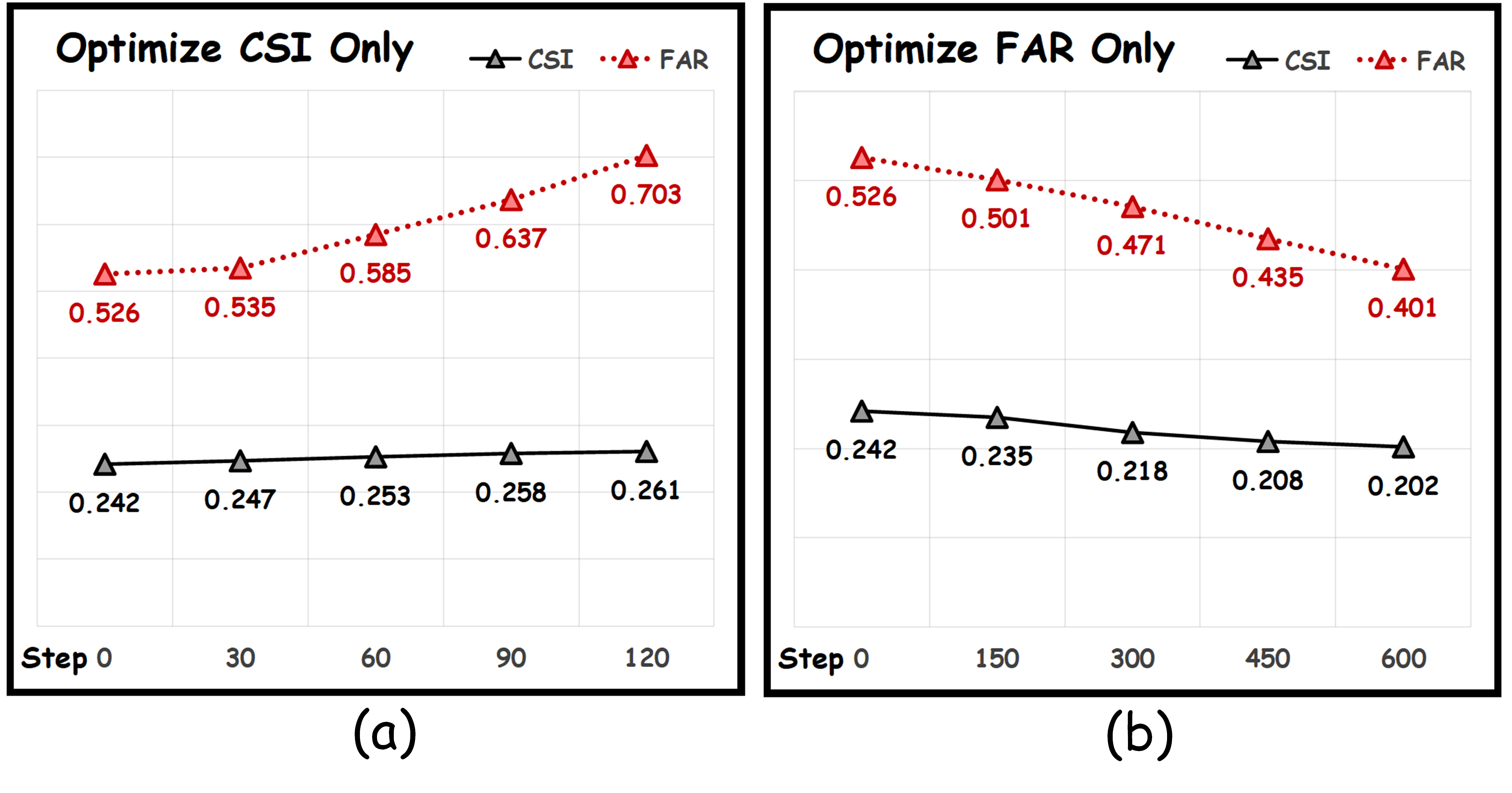}
    \caption{Trade-off between CSI and FAR. When CSI is used as the sole preference target, the model consistently improves CSI but at the cost of increased FAR, and vice versa.}
    \label{fig:Diff_2}
\end{figure}

However, in practice, real-world evaluation metrics are often multidimensional~\cite{yang2021spatiotemporal, wen2018generating, wu2023discovqa}. 
In fields such as image and video generation, as well as natural language processing, generated results need to satisfy multiple quality and style criteria simultaneously~\cite{min2025towards, hartwig2025survey, xie2024video, you2024towards, soni2025evaluating, lin2023llm}. 
Compared to these domains, evaluations in scientific fields such as precipitation nowcasting are more complex, involving not only a greater number of metrics, but also some that may conflict with one another~\cite{tan2023temporal, ritvanen2025cell}.
Specifically, such metrics include CSI and FAR, which the former emphasizes the hit rate of precipitation events, while the latter measures the proportion of false positives.
Intuitively, making the model more sensitive to precipitation events can improve CSI but often increases false alarms, raising FAR; conversely, prioritizing the reduction of false alarms may cause the model to miss true precipitation events, thereby lowering CSI.
Previous studies~\cite{luo2022experimental} have also shown that many methods improve CSI at the cost of increasing FAR. 
In addition, our preliminary experimental results in Fig.~\ref{fig:Diff_2} further confirm this trade-off: Using CSI as the preference signal enables continuous improvement in CSI, but this frequently leads to a deterioration in FAR, and vice versa.
This raises a key question: \textbf{\textit{How Can We Jointly Optimize Conflicting Metrics CSI and FAR to Reach Their Pareto Frontier?}}

To address this, we propose SynCast, a model for precipitation nowcasting that uses the Diffusion Sequential Preference Optimization (Diffusion-SPO) post-training framework to synergize conflicting metrics.
Inspired by the painting process of ``first removing excess, then refining the details”, Diffusion-SPO adopts a two-stage preference alignment strategy. 
In the first stage, it focuses on reducing FAR, learning to suppress excessive false alarms. And in the second stage, it aims to improve CSI while maintaining the FAR alignment by incorporating a corresponding constraint into the objective function. 
We evaluate SynCast on three radar precipitation datasets. Experimental results show that SynCast is able to reduce FAR while simultaneously improving CSI and achieves performance comparable to the current state-of-the-art models.

Our main contributions can be summarized as:
\begin{itemize}
    \item For probabilistic generative models in precipitation nowcasting, we first propose a post-training framework based on reinforcement learning to stably optimize key forecasting metrics, consistently achieving the best performance.
    \item We demonstrate the conflict between CSI and FAR in nowcasting and introduce SynCast, which employs Diffusion Sequential Preference Optimization to simultaneously improve these conflicting metrics.
    \item We validate SynCast across different datasets and evaluate the Diffusion-SPO framework across various diffusion models, demonstrating the strong generalization and effectiveness of both the model and the framework.
\end{itemize}

The rest of the paper is organized and detailed below. Section II reviews related work on precipitation nowcasting and preference-based optimization methods. Section III introduces the task definitions and relevant preliminaries. Section IV details the training of our proposed SynCast model. Section V presents experimental results, demonstrating the effectiveness of SynCast and the Diffusion-SPO framework across multiple datasets and models. Finally, Section VI concludes the paper and discusses future directions.

\section{Related Works}
\subsection{Deep Learning-based Precipitation Nowcasting}
Existing approaches for precipitation nowcasting can be broadly divided into two categories: deterministic and probabilistic. 
Deterministic models generally aim to minimize the mean squared error between predicted and target sequences
For example, ConvLSTM~\cite{shi2015convolutional} incorporates convolutional operations into LSTM to simultaneously capture both spatial and temporal patterns. PredRNN~\cite{wang2022predrnn} models multi-level spatiotemporal dependencies via decoupled memory cells and a zigzag memory flow. 
And EarthFormer~\cite{gao2022earthformer} employs cuboid attention to efficiently capture both intra-frame and inter-frame temporal relationships. 
However, these methods generally suffer from blurry predictions.
In contrast, probabilistic approaches model the distribution of future sequences instead of their expected. PreDiff~\cite{gao2023prediff} leverages conditional latent diffusion combined with explicit knowledge alignment to achieve physically consistent probabilistic spatiotemporal forecasts.
DiffCast~\cite{yu2024diffcast} decomposes the precipitation system into global deterministic motion and local stochastic variations, and uses a diffusion model to capture local uncertainty.
Building on DiffCast, AFGDiff~\cite{ji2025afgdiff} introduces a generative adversarial network as a closed-loop feedback to continuously evaluate and optimize the model. Although probabilistic methods are effective in capturing local details and extreme events, their inherent stochasticity often prevents them from consistently approaching the upper bound of performance metrics.

\subsection{Preference Optimization in Generative Models}
Preference optimization methods in generative models mainly follow three technical routes~\cite{wu2025preference}. 
Initial work was dominated by reinforcement learning and policy optimization approaches, in which a reward model is trained to learn human preferences and subsequently used with algorithms such as PPO~\cite{schulman2017proximal} to maximize expected rewards under KL regularization. 
For diffusion models, DDPO~\cite{black2023training} treats denoising as a multi-step decision process and applies policy gradients to optimize downstream rewards, but this comes with high computational cost. 
In contrast, DPO~\cite{rafailov2023direct} directly optimizes on human preference pairs using a binary classification objective. 
It links the preference loss to the optimal policy via a closed-form solution, eliminating the need for an explicit reward model and avoiding the instability and cost of RL. 
Diffusion-DPO~\cite{wallace2024diffusion} extends this approach to diffusion models by integrating it with the denoising process.
The third approach leverages differentiable reward models for end-to-end fine-tuning. 
ReFL~\cite{xu2023imagereward} uses ImageReward as supervision and achieves strong performance on both human evaluations and automatic metrics. 
DRaFT~\cite{clark2023directly} and AlignProp~\cite{prabhudesai2023aligning} directly backpropagate the reward gradients through the diffusion model, with AlignProp further employing low-rank adapters and gradient checkpoint to reduce memory usage.
However, existing methods have been primarily applied to text-to-image generation, and their effectiveness for precipitation nowcasting tasks has not yet been explored.

\section{Task Definition and Preliminaries}
\subsection{Task Definition} 
Precipitation nowcasting is typically formulated as a spatiotemporal prediction problem~\cite{shi2015convolutional}, similar to video prediction in computer vision~\cite{sheng2024spatial, li2020video, wang2021neural, li2022order}. 
Given the first $T'$ radar frames as input $x \in \mathbb{R}^{T' \times C \times H \times W}$, the task is to predict the following $T-T'$ frames $y \in \mathbb{R}^{(T-T') \times C \times H \times W}$, where $C=1$ denotes the number of channels in each frame, and $H$ and $W$ are the spatial height and width.

\subsection{Preliminaries}
\subsubsection{Diffusion Model}
The Denoising Diffusion Probabilistic Model (DDPM) is a class of generative models that estimates the data distribution. 
It trains a network to iteratively denoise samples corrupted by a Markovian noise process, gradually transforming pure Gaussian noise into realistic samples.
Its generation process consists of two steps: forward noising and reverse denoising. 
In the forward process, the target data $y_0$ is progressively corrupted by Gaussian noise.
After $t$ steps of adding noise, the data becomes $y_t$ and follows the distribution:
\begin{equation}
q(y_t \mid y_0) = \mathcal{N}\big(y_t; \overline{\alpha}_t\, y_0, (1-\overline{\alpha}_t) I\big),
\end{equation}
where $\bar{\alpha}_t = \prod_{s=1}^{t} (1-\beta_s)$ and $\beta_s \in (0, 1)$ is a pre-defined noise schedule. After $T$ diffusion steps, $y_T$ is approximately distributed as standard Gaussian.
In the reverse process, the model aims to generate data by inverting the forward process. This is formalized as learning a reverse-time Markov chain:
\begin{equation}
p_{\theta}(y_{t-1}\mid y_{t}) = \mathcal{N}\big(y_{t-1}; \mu_{\theta}(y_{t},t), \sigma_t^2 I\big),
\label{eq:diff1}
\end{equation}
where $\mu_{\theta}(y_{t},t)$ is the conditional mean function, and it is parameterized as:
\begin{equation}
\mu_{\theta}(y_{t},t) = \frac{1}{\sqrt{\alpha_t}} \Big(y_t - \frac{\beta_t}{\sqrt{1-\bar{\alpha}_t}} \, \epsilon_{\theta}(y_t,t)\Big),
\label{eq:diff2}
\end{equation}
where $\epsilon_{\theta}(y_t,t)$ is a denoise neural network that predicts the noise added to $y_t$ during the forward diffusion process. The training objective is to minimize the difference between the predicted and true noise, with the following loss function:
\begin{equation}
\mathcal{L}(\theta) = \mathbb{E}_{y_0, t, \epsilon} \Big[ \|\epsilon - \epsilon_{\theta}(y_t,t)\|^2 \Big].
\end{equation}
During the reference stage, samples are progressively refined from pure noise following the diffusion reverse process:
\begin{equation}
y_{t-1} = \frac{1}{\sqrt{\alpha_t}} \Big(y_t - \frac{\beta_t}{\sqrt{1-\bar{\alpha}_t}} \, \epsilon_{\theta}(y_t,t)\Big) + \sigma_t \epsilon, \quad \epsilon \sim \mathcal{N}(0,I),
\end{equation}
where $\sigma_t$ is usually set to $\sqrt{\beta_t}$, and the noise term $\epsilon$ ensures diversity in the generated samples.

\subsubsection{Diffusion-DPO}
Diffusion-DPO is a preference-based fine-tuning framework that extends the DPO approach to diffusion models. 
In standard DPO, preference alignment is achieved without explicitly fitting a reward model. 
Instead, the target model is trained to increase the relative likelihood of preferred samples compared to a reference model. The corresponding training objective is formulated as:
\begin{equation}
\begin{aligned}
\mathcal{L}_{\text{DPO}}(\theta) 
= - \mathbb{E}_{(x, y_0^w, y_0^l) \sim D} 
\Bigg[ \log \sigma \Bigg( \beta \Big( 
\log \frac{p_\theta(y_0^w \mid x)}{p_{\text{ref}}(y_0^w \mid x)} \\
- \log \frac{p_\theta(y_0^l \mid x)}{p_{\text{ref}}(y_0^l \mid x)} 
\Big) \Bigg) \Bigg],
\end{aligned}
\end{equation}
where $p_{\text{ref}}$ denotes the frozen reference distribution, which is typically defined by a supervised fine-tuned model. 
$x$ represents the prompt, while $y_w$ and $y_l$ are the preferred and dispreferred responses to $x$, respectively.
$\beta$ is a hyperparameter controlling the strength of the preference alignment.
\begin{figure*}[ht]
    \centering
    \includegraphics[width=0.9\linewidth]{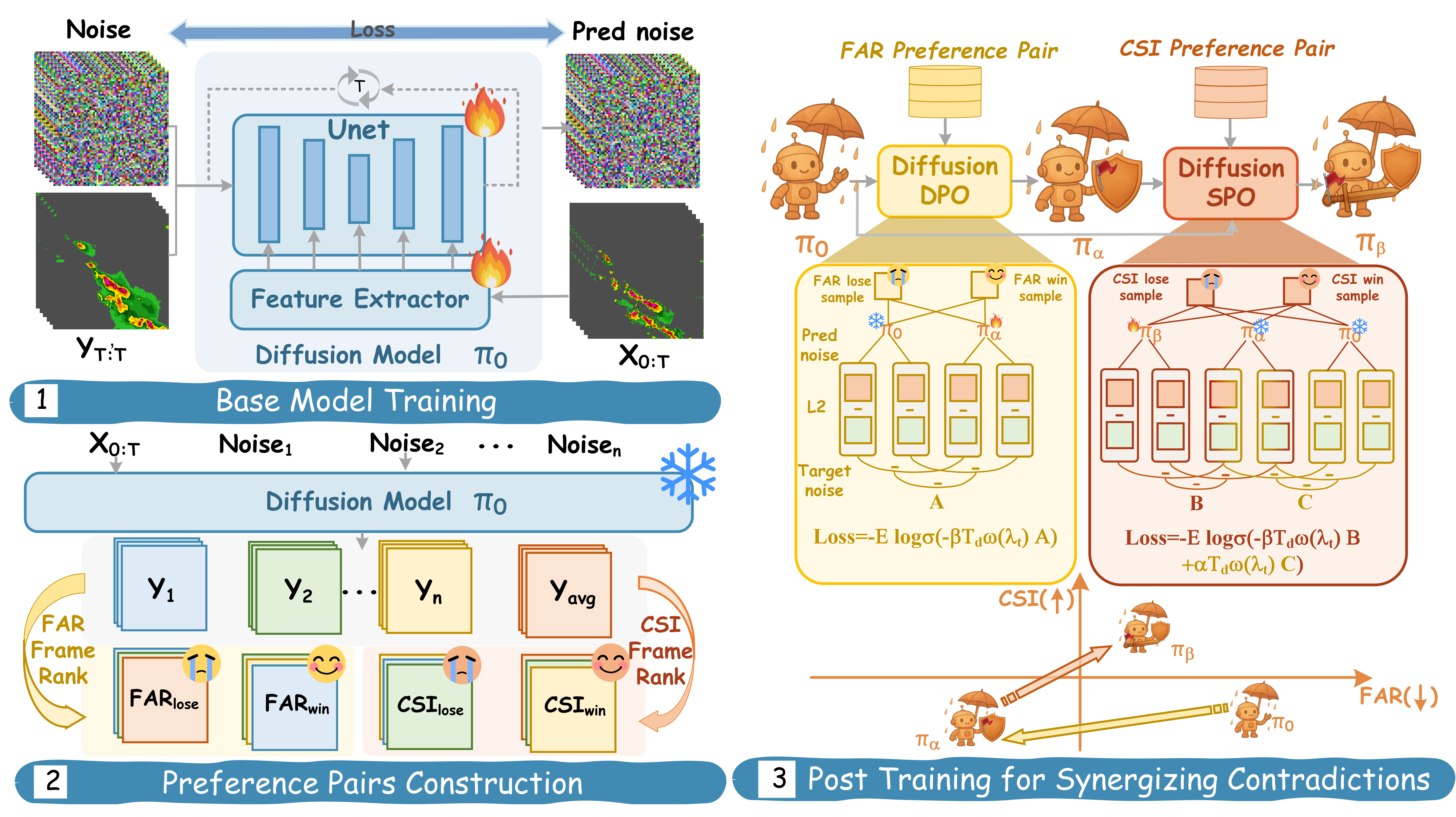}
    \caption{The workflow of SynCast. The training of SynCast consists of three steps. First, a UNet-based diffusion model is trained in pixel space, with conditional information injected via a CNN-based feature extractor to facilitate denoising. Next, we construct preference pairs for conflicting metrics. Specifically, by varying the initial noise, we generate $n$ predictions for each input, along with their average, resulting in $n+1$ candidates. For each predicted frame, FAR and CSI are computed. The best frame at each lead time is treated as a win sample, while the worst is regarded as a lose sample. Finally, reinforcement fine-tuning is performed. We first use Diffusion-DPO to align the model with respect to FAR, and then apply Diffusion-SPO to optimize CSI under the constraint of preserving the FAR alignment achieved in the previous stage.}
    \label{fig:frameflow}
\end{figure*}

However, for the diffusion model, directly optimizing the target distribution $p_\theta(y_0|x)$ is infeasible. As it requires marginalizing over all possible diffusion paths $y_{1:T}$ that could generate $y_0$. 
To overcome this, Diffusion-DPO introduces latent variables $y_{1:T}$ and leverages the evidence lower bound (ELBO) to approximate the marginal.
It further defines a reward function over the entire diffusion path $R(x, y_{0:T})$, and considers the expected reward for the final sample as:
\begin{equation}
r(x, y_0) = \mathbb{E}_{p_\theta(y_{1:T} \mid y_0, x)} \big[ R(x, y_{0:T}) \big].
\end{equation}
For the KL regularization term, it adopts a pathwise joint KL upper bound:
\begin{equation}
\text{KL}\big[ p_\theta(y_{0:T} \mid x) \parallel p_{\text{ref}}(y_{0:T} \mid x) \big].
\end{equation}
By combining the reward and KL terms, the pathwise Diffusion-DPO optimization objective can be expressed as:
\begin{equation}
\begin{split}
& \hspace{-5mm}\max_{p_\theta} \mathbb{E}_{x \sim D, y_{0:T} \sim p_\theta(y_{0:T}|x)} \big[ r(x, y_0) \big] - \\
& \hspace{10mm}\beta \, \text{KL}\big[ p_\theta(y_{0:T} \mid x) \parallel p_{\text{ref}}(y_{0:T} \mid x) \big].  
\end{split}
\end{equation}
Following a similar process as in the standard DPO objective, the expectation is expanded over the diffusion trajectories to define a concrete pathwise objective:
\begin{equation}
\begin{split}
& \hspace{-0mm} \mathcal{L}(\theta) = - \mathbb{E}_{(y_0^w, y_0^l) \sim D} 
\log \sigma \Bigg(
\beta \, \mathbb{E}_{\substack{y_{1:T}^w \sim p_\theta(y_{1:T}^w \mid y_0^w),\, \\ y_{1:T}^l \sim p_\theta(y_{1:T}^l \mid y_0^l)}}
\\
& \hspace{-0mm}\Big[
\log \frac{p_\theta(y_{0:T}^w)}{p_{\text{ref}}(y_{0:T}^w)} - \log \frac{p_\theta(y_{0:T}^l)}{p_{\text{ref}}(y_{0:T}^l)}.
\Big]
\Bigg)
\end{split}
\end{equation}
To make the objective tractable, Diffusion-DPO applies two approximations: first, Jensen's inequality, and second, replacing expectations over the reverse process $p_\theta(y_{1:T} \mid y_0)$ with over the forward process $q(y_{1:T} \mid y_0)$. The resulting loss can then be written as:
\begin{equation}
\begin{split}
\mathcal{L}(\theta) = - \mathbb{E}_{(y^w_0, y^w_0) \sim D, t \sim U(0,T), y_t^w \sim q(y_t^w \mid y_0^l), y_t^l \sim q(y_t^l \mid y_0^l)}\\
& \hspace{-50mm} \log \sigma \Big(- \beta T \, \omega(\lambda_t)\big( \\
& \hspace{-60mm} \| \epsilon^w - \epsilon_\theta(y_t^w, t) \|_2^2 - \| \epsilon^w - \epsilon_{\text{ref}}(y_t^w, t) \|_2^2 \\
-( \| \epsilon^w - \epsilon_\theta(y_t^l, t) \|_2^2 - \| \epsilon^w - \epsilon_{\text{ref}}(y_t^l, t) \|_2^2 \big) \Big),  
\end{split}
\end{equation}
where $\omega(\lambda_t)$ is a timestep-dependent weighting factor. 

\section{Our Framework: SynCast}
The training of our SynCast consists of three steps: 
(i) training the probabilistic base model $\pi_0$;
(ii) constructing FAR and CSI preference pairs by evaluating the corresponding metrics of candidate samples generated from $\pi_0$; and 
(iii) performing Diffusion-SPO for post-training. 
It is worth noting that the effectiveness of step (iii) relies on the diversity of preference samples, which is mainly from the stochasticity of the $\pi_0$ in step (i), and partly from the level of granularity in the sample filtering process in step (ii). 
Through steps (i)–(iii), SynCast effectively reduces the FAR while improving the CSI in probabilistic precipitation nowcasting tasks, achieving joint optimization of these conflicting metrics. The overall workflow of SynCast is illustrated in Fig.~\ref{fig:frameflow}.

\subsection{Probabilistic Base Model}
To provide sufficient sample diversity for preference optimization, we design the base model to enhance stochasticity as much as possible.
Unlike DiffCast~\cite{yu2024diffcast} and CasCast~\cite{gong2024cascast}, which incorporate deterministic branches in their architectures, SynCast adopts a fully probabilistic formulation. 
This avoids the limitation of deterministic components on the generative space, as their fixed mappings override random perturbations and thereby suppress stochastic variations in the sampling process.
Moreover, we train the base model directly in the pixel space. Compared with the latent diffusion models, our pixel-space model preserves the high resolution of the original data and offers a larger, more flexible sampling space. 
For the network architecture, we adopt a UNet-based denoising network, in which conditional input frames are first encoded by a convolutional neural network (CNN) based feature extractor with residual connections. 
The extracted features are subsequently injected into the UNet through attention mechanisms to provide contextual guidance.

\subsection{Construction of Preference Pairs}
To enrich the diversity of preference pairs, we adopt a frame-wise construction strategy. 
Given an input~$x$, we first initialize~$N$ independent noise sequences and progressively denoise them into candidate sequences~$\{y^i_0\}_{i=1}^N$. 
These $N$ sequences are then ensembled into an extra candidate, resulting in a total of $N+1$ candidate sequences.
Instead of keeping or discarding entire sequences, we evaluate every frame. 
For each leading time, we rank $N+1$ frames by $\text{FAR}$ and $\text{CSI}$, extract the best and worst frames for each metric.
These frames are reassembled into win–lose samples, forming two preference pairs: $\langle y_w^{\text{CSI}}, y_l^{\text{CSI}}\rangle$ and $\langle y_w^{\text{FAR}}, y_l^{\text{FAR}}\rangle$.
This frame-wise approach could preserve fine-grained intra-sequence variations and supply the optimizer with sharper supervisory gradients, leading to more effective preference learning.

\subsection{Diffusion-SPO for Joint Optimization of Conflicting Metrics}
The denominator of FAR contains only hits and false alarms, making it highly sensitive to a few false detections, especially under weak or no precipitation conditions. Unlike FAR, CSI jointly considers hits, misses, and false alarms, resulting in more stable performance across samples. Results on Fig.~\ref{fig:Diff} confirm that FAR fluctuates much more than CSI.
Directly optimizing CSI drives the model to expand prediction regions for higher hit rates, but this also increases false alarms, resulting in a sharp rise in FAR and unstable overall performance.

To avoid instability, our proposed Diffusion-SPO adopts an iterative strategy conceptually similar to the process of painting: ``first removing excess, then refining the details''. 
Diffusion-SPO begins by using FAR as an eraser, wiping away the model’s over-predictions. 
After that, it swaps the eraser for a fine brush, focusing on CSI to refine the places that truly matter, while imposing constraints to preserve the results achieved during the FAR stage.

Specifically, the first phase is carried out without any constraints.
We therefore directly apply Diffusion-DPO~\cite{wallace2024diffusion} to base model $\pi_0$, obtaining FAR-aligned policy $\pi_{\alpha}$, which maximizes the FAR preference reward:

\begin{equation}
r_{\text{FAR}}(x, y_0) = \mathbb{E}_{y_{1:T_d} \sim \pi_{\alpha}(y_{1:T_d} \mid y_0, x)} \big[ R_{\text{FAR}}(x, y_{0:T_d}) \big],
\end{equation}
where $T_d$ denotes the diffusion timestep, $y_0$ is the clean prediction, and $R_{\text{FAR}}(x, y_{0:T_d})$ denotes the FAR preference reward for a diffusion trajectory generating prediction $y_0$.

In the second phase, we focus on aligning with CSI preferences while introducing an additional constraint to prevent FAR alignment performance. 
This results in the following constrained optimization problem:
\begin{equation}
\begin{aligned}
& \max_{\pi_{\beta}} \; 
\mathbb{E}_{x \sim D_x, y_{0:T_d} \sim \pi_{\beta}(x, y_{0:T_d})}
\big[ r_{\text{CSI}}(x, y_0) \big] \\[6pt]
 \text{s.t.} \quad 
& D_{\mathrm{KL}}\!\left(
\pi_{\beta}(y_{0:T_d} \mid x) \,\big\|\, \pi_{\alpha}(y_{0:T_d} \mid x)
\right) \leq H_0 \\[6pt]
& 
\mathbb{E}_{x \sim D_x, y_{0:T_d} \sim \pi_{\beta}(y_{0:T_d} \mid x)}
\big[r_{\text{FAR}}(x, y_0) \big] \geq H_1,
\end{aligned}
\label{eq:original}
\end{equation}
where $D_x$ denotes the dataset of input radar sequences, $\pi_{\beta}$ denotes the policy aligned with the CSI preference, $H_0$ and $H_1$ are the thresholds for the KL-divergence constraint and the FAR preference reward, respectively. 
$r_{\text{CSI}}(x, y_0) = \mathbb{E}_{y_{1:T_d} \sim \pi_{\beta}(y_{1:T_d} \mid y_0, x)} \big[ R_{\text{CSI}}(x, y_{0:T_d}) \big]$ is the expected CSI preference reward of $y_0$ over all its generating paths. This formulation ensures three key goals: 
(i) maximizing the CSI preference reward, 
(ii) constraining deviations from the $\pi_{\alpha}$, and 
(iii) preventing excessive degradation of FAR preference alignment. 
However, directly solving Eq.~\eqref{eq:original} is challenging, so we convert it into its Lagrangian dual form:
\begin{equation}
\begin{aligned}
arg\max_{\pi_{\beta}} \Bigg\{ 
& \mathbb{E}_{x \sim D_x, y_{0:T_d} \sim \pi_{\beta}} \Big[ r_{\text{CSI}}(x, y_0) + \alpha r_{\text{FAR}}(x, y_0) \Big] \\
& - \beta D_{\mathrm{KL}}\Big( \pi_{\beta}(y_{0:t} \mid x) \;\|\; \pi_{\alpha}(y_{0:T_d} \mid x) \Big)
\Bigg\},
\end{aligned}
\label{eq: Lag}
\end{equation}
where $\alpha\ge 0$ controls the influence of FAR preference rewards $R_{FAR}$, and $\beta \ge 0$ is the coefficient for the KL constraint. 
Like prior work~\cite{lou2024spo}, the policy $\pi_{\beta}(y_{0:T_d}|x)$ has a closed-form solution in Eq.~\eqref{eq: Lag} (The detailed derivation is provided in Supplementary Material A) :
\begin{equation}
\begin{aligned}
\pi^*_{\beta}(y_{0:T_d} \mid x) 
= & \frac{1}{Z_\text{CSI}(x)} \, 
\pi_{\alpha}(y_{0:T_d} \mid x) \, \\
& \exp\Bigg[
    \frac{1}{\beta} \Big(
        r_\text{CSI}(x, y_0) + \alpha r_{\text{FAR}}(x, y_0)
    \Big)
\Bigg],
\end{aligned}
\label{eq:closed-form}
\end{equation}
where $Z_\text{CSI}(x)=\sum_{y_0} \pi_{\alpha}(y_0\mid x) \exp\Big( \frac{\alpha}{\beta} r_\text{FAR}(x, y_0) + \frac{1}{\beta} r_\text{CSI}(x, y_0) \Big)$ is the partition function that ensures normalization. By taking the logarithm of Eq.~\eqref{eq:closed-form}, the CSI preference reward can be expressed as:
\begin{equation}
\begin{aligned}
r_\text{CSI}(x, y_0) 
&= - \alpha \, r_\text{FAR}(x, y_0) \\
&\quad + \beta \log \frac{\pi_{\beta}(y_{0:T_d}\mid x)}{\pi_{\alpha}(y_{0:T_d}\mid x)} 
+ \beta \log Z_{\text{CSI}}(x),
\end{aligned}
\end{equation}
where $r_{\text{FAR}}(x, y_0) = \beta \log \frac{\pi_{\alpha}(y_{0:t} \mid x)}{\pi_0(y_{0:T_d} \mid x)} 
+ \beta \log Z_{\text{FAR}}(x)$ also be represented by the $\pi_0$ and $\pi_{\alpha}$.

However, CSI and FAR metrics cannot be used directly as rewards, since they are non-differentiable. 
Instead, a Bradley-Terry (BT) model is used, in which preference is determined by the difference in rewards between trajectories: $P(y_{w}^{\text{CSI}} \succ y_{l}^{\text{CSI}} \mid x) = \sigma\!\left( r_{\text{CSI}}(x, y_{w}^{\text{CSI}}) - r_{\text{CSI}}(x, y_{l}^{\text{CSI}}) \right)$, where $\sigma(x) = \frac{1}{1 + e^{-x}}$ is the sigmoid function. 
Incorporating $r_{\text{CSI}}$ into the BT model yields a differentiable preference likelihood, and $\pi_{\beta}$ is optimized by minimizing the negative log-likelihood of the preference pairs as follows:
{\small
\begin{equation}
\begin{aligned}
& \mathcal{L}_{\text{Diffusion-SPO}}{(\pi_{\beta})}= - \mathbb{E}_{(y_0^w, y_0^l) \sim D_p^\text{CSI}}
\log \sigma \\\Bigg( 
& \beta \, \mathbb{E}_{\substack{y_{1:T_d}^w \sim \pi_{\beta}(\cdot \mid y_0^w) \\ 
                               y_{1:T_d}^l \sim \pi_{\beta}(\cdot \mid y_0^l)}}  
\Big[ \log \tfrac{\pi_{\beta}(y_{0:T_d}^w \mid x)}{\pi_{\alpha}(y_{0:T_d}^w \mid x)}
   - \log \tfrac{\pi_{\beta}(y_{0:T_d}^l \mid x)}{\pi_{\alpha}(y_{0:T_d}^l \mid x)} \Big] \\
& - \alpha \beta \, \mathbb{E}_{\substack{y_{1:T_d}^w \sim \pi_{\beta}(\cdot \mid y_0^w) \\ 
                                         y_{1:T_d}^l \sim \pi_{\beta}(\cdot \mid y_0^l)}}
\Big[ \log \tfrac{\pi_{\alpha}(y_{0:T_d}^w \mid x)}{\pi_0(y_{0:T_d}^w \mid x)}
   - \log \tfrac{\pi_{\alpha}(y_{0:T_d}^l \mid x)}{\pi_0(y_{0:T_d}^l \mid x)} \Big]
\Bigg),
\end{aligned}
\label{eq:spo1}
\end{equation}
}
where $D_p^\text{CSI}$ denotes the dataset of CSI preference pairs. 

Following~\cite{wallace2024diffusion}, by applying Jensen’s inequality, replacing the reverse process with its forward counterpart, and utilizing the ELBO, this loss can be reformulated as a tractable, differentiable denoising-style objective (The detailed derivation is provided in Supplementary Material B):
{\small
\begin{equation}
\begin{aligned}
& \mathcal{L}_{\text{Diffusion-SPO}}(\pi_{\beta})^*
= -\mathbb{E}_{\substack{y_0^w, y_0^l \sim D_p^{\text{CSI}}, t \sim \mathcal{U}(0,T_d), \\ 
y_t^w \sim q(y_t^w \mid y_0^w), y_t^l \sim q(y_t^l \mid y_0^l)}} \log \sigma \\ & \Bigg( - \beta T_d \, \omega(\lambda_t) \, 
\Big[
\| \epsilon^w - \epsilon_\text{CSI}(y_t^w, t, x) \|_2^2 
- \| \epsilon^w - \epsilon_{{\text{FAR}}}(y_t^w, t, x) \|_2^2 
\\ & - \| \epsilon^l - \epsilon_\text{CSI}(y_t^l, t, x) \|_2^2 
+ \| \epsilon^l - \epsilon_{\text{FAR}}(y_t^l, t, x) \|_2^2
 \Big] \\ & + \alpha \beta T_d \, \omega(\lambda_t)  \Big[
\| \epsilon^w - \epsilon_\text{FAR}(y_t^w, t, x) \|_2^2 
- \| \epsilon^w - \epsilon_{\pi_0}(y_t^w, t, x) \|_2^2 
\\ & - \| \epsilon^l - \epsilon_\text{FAR}(y_t^l, t, x) \|_2^2 
+ \| \epsilon^l - \epsilon_{\pi_0}(y_t^l, t, x) \|_2^2
\Big]
 \Bigg) \, ,
\end{aligned}
\label{eq:spo2}
\end{equation}}
where $y_t^* = \alpha_t y_0^* + (1-\alpha_t)\epsilon^*$, $\epsilon \sim \mathcal{N}(0,I)$, 
$\lambda_t = \tfrac{\alpha_t^2}{1-\alpha_t^2}$ is the signal-to-noise ratio, and $\omega(\lambda_t)$ is a weighting function. 
The Diffusion-SPO loss not only maximizes the preference reward for CSI, but also preserves the alignment performance of FAR, preventing the model from degrading in FAR excessively when optimizing solely for CSI.
\begin{figure*}[ht]
    \centering
    \includegraphics[width=0.75\linewidth]{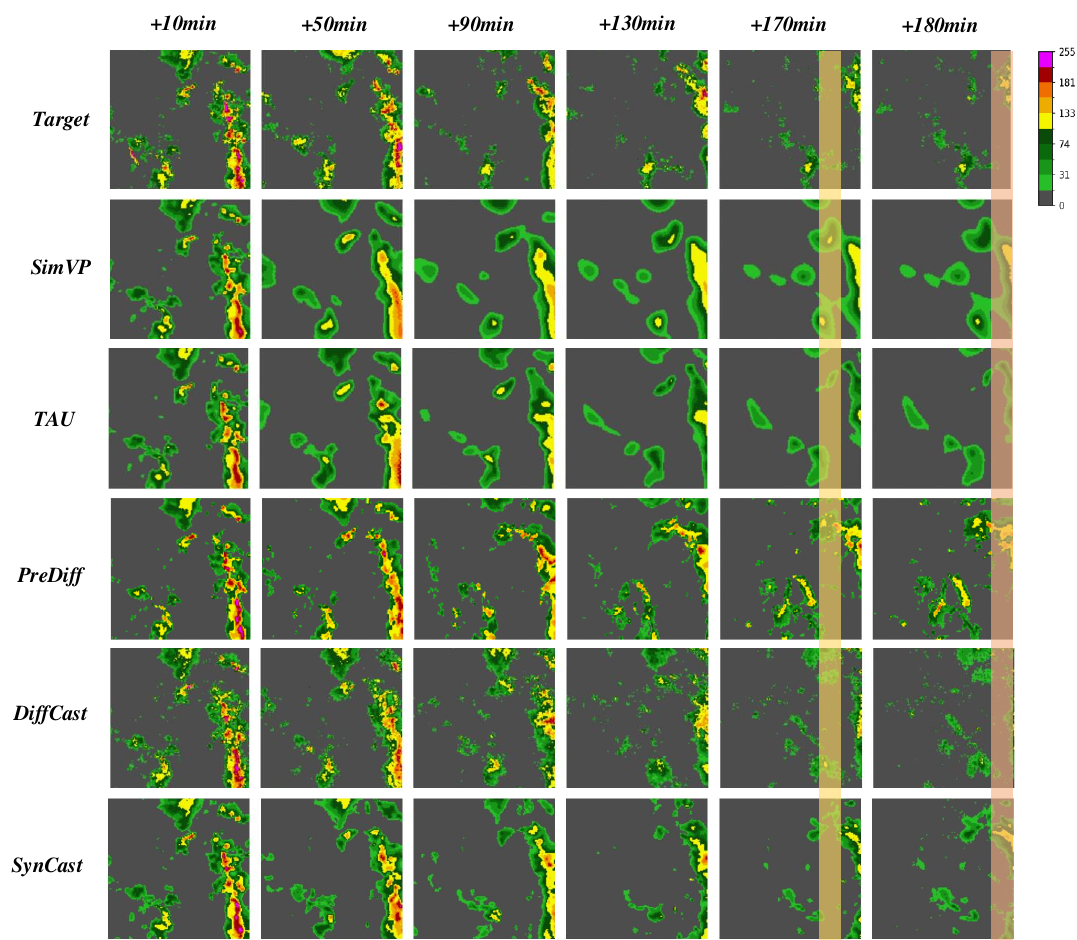}
    \caption{Visual comparison of a set of SEVIR precipitation events. Columns from left to right represent lead times from 10 to 180 minutes, and rows from top to bottom show Target, SimVP, TAU, PreDiff, DiffCast, and SynCast predictions. The yellow-marked region highlights excessive false alarms produced by PreDiff, while the red-marked region indicates a medium-intensity precipitation patch missed by DiffCast.}
    \label{fig:main}
\end{figure*}

\begin{figure*}[ht]
    \centering
    \includegraphics[width=1\linewidth]{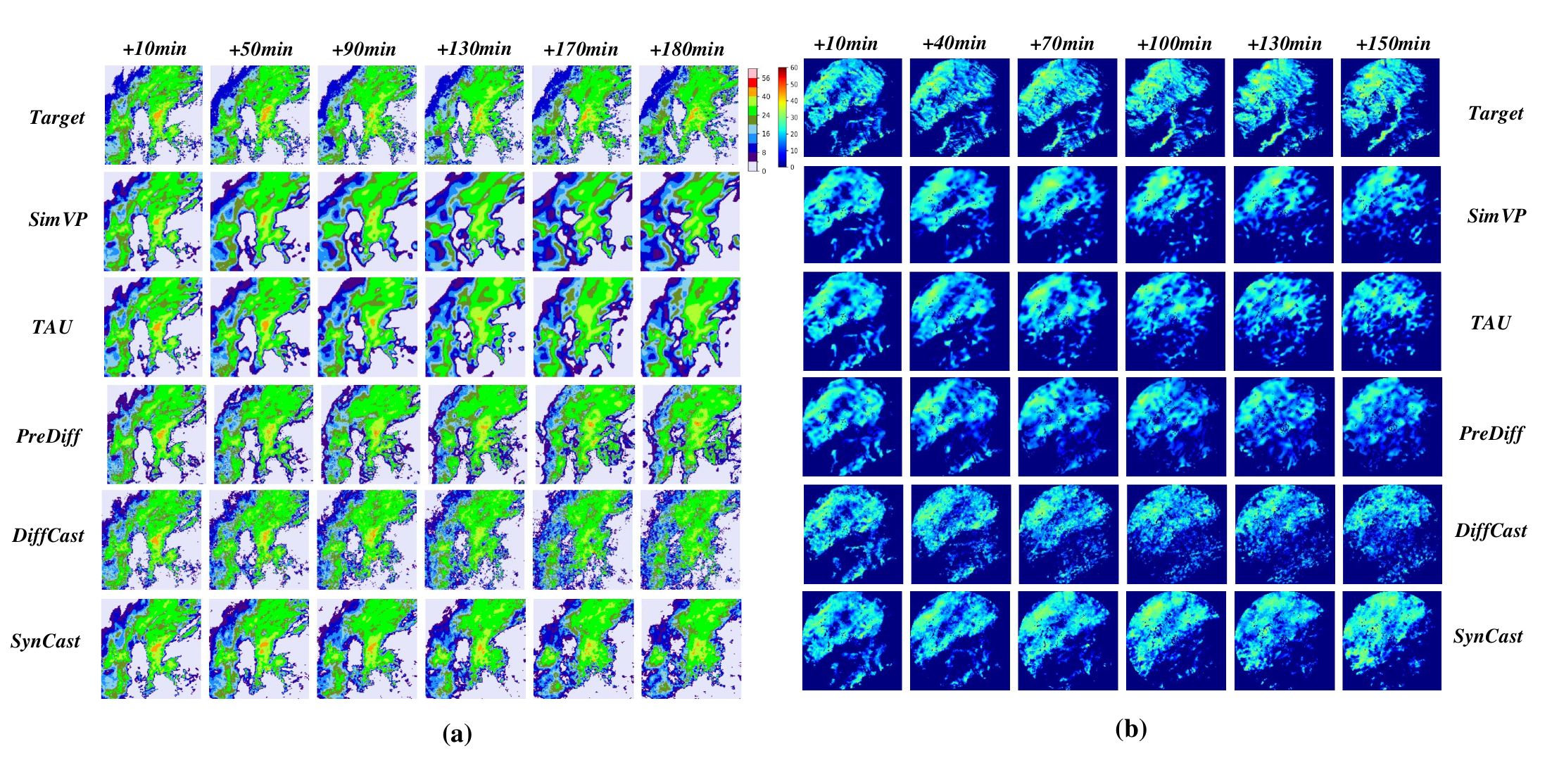}
    \caption{Visual comparison of precipitation events: (a) MeteoNet and (b) HKO-7. }
    \label{fig:union}
\end{figure*}

\begin{table*}
\centering
\caption{Performance comparison of SynCast against deterministic and probabilistic models on the SEVIR dataset. CSI-M denotes the mean CSI across thresholds [16, 74, 133, 160, 181, 219]. POOL1 corresponds to CSI at the original grid resolution, while POOL4 and POOL16 indicate 4×4 and 16×16 grid aggregations, respectively. Best values are highlighted in bold.}
\label{tab:sevir_results}
\begin{tabular}{ccccccccccccc}
\toprule
\multirow{2}{*}{Model} & \multirow{2}{*}{CRPS$\downarrow$} & \multirow{2}{*}{HSS $\uparrow$} & \multicolumn{3}{c}{CSI-M $\uparrow$} & \multicolumn{3}{c}{CSI-74 $\uparrow$} & \multicolumn{3}{c}{CSI-131 $\uparrow$} \\
\cmidrule(r){4-6} \cmidrule(r){7-9} \cmidrule(r){10-12}
 & & & POOL1 & POOL4 & POOL16 & POOL1 & POOL4 & POOL16 & POOL1 & POOL4 & POOL16 \\
\midrule
PredRNN~\cite{wang2022predrnn}     & 0.0480 & 0.290 & 0.219 & 0.241 & 0.257 & 0.448 & 0.473 & 0.529 & 0.147 & 0.173 & 0.194 \\
SimVP~\cite{gao2022simvp}       & 0.0462 & 0.318 & 0.238 & 0.269 & 0.294 & 0.469 & 0.499 & 0.568 & 0.179 & 0.216 & 0.273 \\
Earthformer~\cite{gao2022earthformer} & 0.0466 & 0.289 & 0.221 & 0.249 & 0.284 & 0.463 & 0.495 & 0.564 & 0.156 & 0.193 & 0.256 \\
TAU~\cite{tan2023temporal}         & 0.0465 & 0.316 & 0.237 & 0.264 & 0.287 & 0.467 & 0.497 & 0.573 & 0.180 & 0.215 & 0.268 \\
\midrule
PreDiff~\cite{gao2023prediff}    & 0.0463 & 0.227 & 0.248 & 0.277 & \textbf{0.306} & 0.452 & 0.487 & 0.564 & 0.195 & 0.228 & 0.290 \\
DiffCast~\cite{yu2024diffcast}    & 0.0464 & 0.318 & 0.235 & 0.266 & 0.299 & 0.436 & 0.478 & 0.564 & 0.164 & 0.179 & 0.251 \\
\midrule
SynCast     & \textbf{0.0460} & \textbf{0.357} & \textbf{0.251} & \textbf{0.282} & 0.305 & \textbf{0.475} & \textbf{0.507} & \textbf{0.577} & \textbf{0.207} & \textbf{0.241} & \textbf{0.293} \\
\bottomrule
\end{tabular}
\end{table*}

\begin{table*}[ht]
\centering
\caption{Performance comparison of SynCast against deterministic and probabilistic models on the MeteoNet and HKO-7 datasets. Best values are highlighted in bold.}
\label{tab:cross_dataset}
\begin{tabular}{ccccccccccc}
\toprule
\multirow{3}{*}{Model} & \multicolumn{5}{c}{MeteoNet} & \multicolumn{5}{c}{HKO-7} \\
\cmidrule(r){2-6} \cmidrule(r){7-11}
 & \multirow{2}{*}{CRPS$\downarrow$} & \multirow{2}{*}{HSS$\uparrow$} & \multicolumn{3}{c}{CSI-M$\uparrow$} & \multirow{2}{*}{CRPS$\downarrow$} & \multirow{2}{*}{HSS$\uparrow$} & \multicolumn{3}{c}{CSI-M$\uparrow$} \\
\cmidrule(r){4-6} \cmidrule(r){9-11}
 & & & POOL1 & POOL4 & POOL16 & & & POOL1 & POOL4 & POOL16 \\
\midrule
PredRNN & 0.0203 & 0.248 & 0.168 & 0.189 & 0.197 & 0.0385 & 0.302 & 0.205 & 0.251 & 0.307 \\
SimVP & 0.0203 & 0.350 & 0.237 & 0.278 & 0.280 & 0.0358 & 0.356 & 0.242 & 0.304 & 0.361 \\
Earthformer & 0.0204 & 0.323 & 0.216 & 0.259 & 0.258 & 0.0359 & 0.323 & 0.217 & 0.272 & 0.324 \\
TAU & 0.0202 & 0.353 & 0.231 & 0.279 & 0.280 & 0.0356 & 0.363 & 0.247 & 0.307 & 0.362 \\
\midrule
PreDiff & 0.0232 & 0.295 & 0.213 & 0.252 & 0.256 & 0.0367 & 0.266 & 0.228 & 0.279 & 0.335 \\
DiffCast & 0.0209 & 0.353 & 0.238 & 0.280 & 0.281 & 0.0361 & 0.354 & 0.239 & 0.301 & 0.354 \\
\midrule
SynCast & \textbf{0.0198} & \textbf{0.362} & \textbf{0.242} & \textbf{0.282} & \textbf{0.284} & \textbf{0.0355} & \textbf{0.382} & \textbf{0.251} & \textbf{0.313} & \textbf{0.363} \\
\bottomrule
\end{tabular}
\end{table*}

\begin{table*}
\scriptsize
\centering
\caption{Performance of three probabilistic models after Diffusion-SPO fine-tuning. SynPreDiff, SynDiffCast, and SynCast denote models fine-tuned on PreDiff, DiffCast, and SynCast\textsubscript{ScoreBase}, respectively.}
\label{tab:cross_model}
\resizebox{\textwidth}{!}{%
\begin{tabular}{ccccccccccccc}
\toprule
\multicolumn{1}{c}{} & \multicolumn{4}{c}{SEVIR} & \multicolumn{4}{c}{MeteoNet} & \multicolumn{4}{c}{HKO-7} \\
\cmidrule(r){2-5} \cmidrule(r){6-9} \cmidrule(r){10-13}
\multicolumn{1}{c}{} & & \multicolumn{3}{c}{CSI-M$\uparrow$} & & \multicolumn{3}{c}{CSI-M$\uparrow$} & & \multicolumn{3}{c}{CSI-M$\uparrow$} \\
\cmidrule(r){3-5} \cmidrule(r){7-9} \cmidrule(r){11-13}
\multirow{-3}{*}{Model} & \multirow{-2}{*}{FAR-M$\downarrow$} & POOL1 & POOL4 & POOL16 & \multirow{-2}{*}{FAR-M$\downarrow$} & POOL1 & POOL4 & POOL16 & \multirow{-2}{*}{FAR-M$\downarrow$} & POOL1 & POOL4 & POOL16 \\
\midrule
PreDiff & 0.590 & 0.248 & 0.277 & 0.306 & 0.591 & 0.213 & 0.252 & 0.256 & 0.482 & 0.228 & 0.279 & 0.335 \\
\multirow{2}{*}{SynPreDiff} & 0.577 & 0.258 & 0.286 & 0.294 & 0.577 & 0.227 & 0.272 & 0.264 & 0.473 & 0.248 & 0.309 & 0.371 \\
 & \textbf{(-5.31\%)} & \textbf{(+4.03\%)} & \textbf{(+3.24\%)} & \textbf{(-3.92\%)} & \textbf{(-2.36\%)} & \textbf{(+6.57\%)} & \textbf{(+7.93\%)} & \textbf{(+7.42\%)} & \textbf{(-1.87\%)} & \textbf{(+8.77\%)} & \textbf{(+10.75\%)} & \textbf{(+10.74\%)} \\
 \midrule
DiffCast & 0.601 & 0.235 & 0.266 & 0.299 & 0.553 & 0.238 & 0.280 & 0.281 & 0.535 & 0.239 & 0.301 & 0.354 \\
\multirow{2}{*}{SynDiffCast} & 0.570 & 0.229 & 0.257 & 0.280 & 0.547 & 0.243 & 0.284 & 0.282 & 0.534 & 0.241 & 0.303 & 0.361 \\
 & \textbf{(-5.16\%)} & (-2.55\%) & (-3.38\%) & (-6.35\%) & \textbf{(-1.08\%)} & \textbf{(+2.10\%)} & \textbf{(+1.42\%)} & \textbf{(+0.35\%)} & \textbf{(-0.18\%)} & \textbf{(+0.83\%)} & \textbf{(+0.66\%)} & \textbf{(+1.97\%)} \\
  \midrule
 $\text{SynCast}_\text{ScoreBase}$ & 0.526 & 0.242 & 0.269 & 0.294 & 0.467 & 0.236 & 0.278 & 0.275 & 0.495 & 0.242 & 0.294 & 0.333 \\
\multirow{2}{*}{SynCast} & 0.505 & 0.251 & 0.282 & 0.305 & 0.461 & 0.242 & 0.282 & 0.284 & 0.491 & 0.251 & 0.313 & 0.363 \\
 & \textbf{(-3.99\%)} & \textbf{(+3.71\%)} & \textbf{(+4.83\%)} & \textbf{(+3.74\%)} & \textbf{(-1.28\%)} & \textbf{(+2.54\%)} & \textbf{(+1.43\%)} & \textbf{(+3.27\%)} & \textbf{(-0.6\%)} & \textbf{(+3.71\%)} & \textbf{(+6.46\%)} & \textbf{(+9.01\%)} \\
\bottomrule
\end{tabular}
}
\end{table*}

\section{Experiments}
\subsection{Experimental Setting}
\subsubsection{Datasets}
We conduct our experiments on three radar precipitation datasets: SEVIR~\cite{veillette2020sevir}, MeteoNet~\cite{larvor2021meteonet}, and HKO-7~\cite{shi2017deep} as follows:
\begin{itemize}
    \item \textbf{SEVIR}~\cite{veillette2020sevir} is an Earth observation dataset for the Contiguous United States (CONUS) with radar data sourced from NEXRAD. It provides sequences of radar mosaics of Vertically Integrated Liquid (VIL) and contains over 10,000 weather events. Each event covers a 384 km × 384 km area with a spatial resolution of 384 × 384. Events from 2017–2018 are used for training, and events from 2019 are used for validation and testing.
    \item \textbf{MeteoNet}~\cite{larvor2021meteonet} provides radar-based precipitation data for France, covering the northwest and southeast regions. To ensure data quality, we crop each observation to retain only the top-left 400 × 400 pixels. Observations from 2016–2017 are used for training, and those from 2018 are used for validation and testing.
    \item \textbf{HKO-7}~\cite{shi2017deep} is a radar precipitation dataset for Hong Kong, providing Constant Altitude Plan Position Indicator (CAPPI) images over a 512 km × 512 km region. Data from 2009–2014 are used for training and validation, and 2015 is used for testing.
\end{itemize}

It should be noted that the study does not follow the commonly used setup in~\cite{gong2024cascast}, which uses 1 hour of radar echo at 5–6 minute intervals to predict the next hour.
Instead, we use 1 hour of data at 10–12 minute intervals to predict the following 3 hours. 
This design is consistent with findings that increasing historical frames brings limited improvement~\cite{zhang2023skilful}, and also enhances efficiency by reducing frequent short-term predictions.
Due to computational resource limitations, all datasets are downscaled to 128 × 128.

\subsubsection{Evaluations}
To evaluate the performance of precipitation nowcasting, we first follow~\cite{luo2022reconstitution, gao2023prediff, yu2024diffcast} to compute the CSI and the Heidke Skill Score (HSS), defined as follows:
\begin{equation}
\text{CSI} = \frac{H}{H + M + F},
\end{equation}
\begin{equation}
\text{HSS} = \frac{2 (H \times CN - M \times F)}{(H + M)(M + CN) + (H + F)(F + CN)},
\end{equation}
where $H$, $M$, $F$ and $CN$ denote the number of hits, misses, false alarms, and correct negatives, respectively. 
These metrics are calculated at the pixel level. 
As these metrics are sensitive to small spatial displacements, we mitigate this issue by applying 4 × 4 and 16 × 16 average pooling to forecasts and observations, following~\cite{gao2022earthformer, yu2024diffcast}.
This procedure effectively blur minor location mismatches, providing a more reasonable assessment of the overall spatial structure of local precipitation.
In addition, we use the Continuous Ranked Probability Score (CRPS) to assess the model’s ability to quantify uncertainty.
It measures the difference between the forecast and observation distributions in continuous space, with smaller values indicating better forecast quality. And in the deterministic case, CRPS degenerates to the Mean Absolute Error (MAE)

\subsubsection{Training Details}
We trained the base model of SynCast for 200K iterations with a batch size of 8. The training was optimized using AdamW with a learning rate of $\alpha = 1 \times 10^{-4}$ and beta coefficients $\beta_1 = 0.9$, $\beta_2 = 0.95$. For the setting of diffusion, the number of training timesteps was set to 1000, and during inference, DDIM sampling was performed with 20 steps.
In the post-training stage, the base model was fine-tuned for 100–300 steps using AdamW with a learning rate of $1 \times 10^{-6}$ and a batch size of 4.

\subsection{Compared to State of the Arts}
To validate the forecasting performance of SynCast, we select four deterministic models and two probabilistic models for comparison, as follows:
\begin{itemize}
    \item \textbf{PredRNN}~\cite{wang2022predrnn} is a deterministic recurrent neural network for spatiotemporal sequence prediction. It models modular visual dynamics with decoupled memory cells and a zigzag memory flow to capture multi-level spatiotemporal features of complex environments.
    
    \item \textbf{SimVP}~\cite{gao2022simvp}, a fully convolutional neural network based video prediction model that consists of encoder, translator, and decoder.
    They are used to extract spatial features, learn temporal evolution, and integrate spatiotemporal information to predict future frames, respectively. 
    It achieves efficient prediction using only CNNs, skip connections, and a standard Mean Squared Error loss.

    \item \textbf{Earthformer}~\cite{gao2022earthformer} performs spatiotemporal Earth system forecasting by dividing the input data into multiple cuboids. It applies cuboid-level self-attention in parallel and connects the cuboids via global vectors, efficiently capturing spatial and temporal dependencies.

    \item \textbf{TAU}~\cite{tan2023temporal} improves spatiotemporal predictive learning by modeling temporal dependencies in a parallelizable way. It decomposes temporal attention into intra-frame statical attention and inter-frame dynamical attention, and integrates these attentions within a temporal module to efficiently capture both frame-level features and frame-to-frame correlations.
        
    \item \textbf{PreDiff}~\cite{gao2023prediff} is a conditional latent diffusion model for probabilistic spatiotemporal forecasting. It incorporates an explicit knowledge alignment mechanism to enforce domain-specific physical constraints. This enables the model to handle uncertainty while generating more physically plausible forecasts.
            
    \item \textbf{DiffCast}~\cite{yu2024diffcast} models precipitation systems as a combination of global deterministic motion and local stochastic variations. The global component is generated by a deterministic model, such as SimVP, while the local residuals are captured using a diffusion framework.
        
\end{itemize}

Table~\ref{tab:sevir_results} presents the quantitative results of SynCast and other models on the SEVIR dataset. From these results, we summarize three key findings:
(i) SynCast achieves the leading performance on CSI-M-POOL1 as well as its multi-scale POOL4 and POOL16, indicating a more accurate capture of local precipitation structures.
Compared with PreDiff, the CSI improvement of SynCast is relatively marginal, but it should be noted that PreDiff achieves gains at the cost of higher FAR as shown in Fig.~\ref{fig:main} and Table~\ref{tab:cross_model}. 
In contrast, SynCast improves CSI while maintaining a lower FAR, achieving a more reliable forecasting ability. 
(ii) In addition, Syncast also demonstrates the highest HSS. Unlike CSI, HSS considers not only hits, misses, and false alarms, but also correct negatives, thereby penalizing false alarms more heavily. 
SynCast achieves an HSS improvement of at least 12.12\% over deterministic models and 57.26\% over the probabilistic model PreDiff.
This achievement primarily benefits from post-training with Diffusion-SPO, which incorporates a constraint during CSI optimization to preserve the previously achieved low FAR. 

This strategy effectively mitigates the trade-off in conventional methods, where increasing CSI often leads to higher FAR, complementing the observations in (i).
(iii) Under the CSI-74/133 metrics, SynCast continues to outperform other models, reflecting its ability to suppress false alarms while still effectively capturing common precipitation events.

Moreover, SynCast exhibits generalization in cross-dataset experiments. As shown in Table~\ref{tab:cross_dataset} and Fig.~\ref{fig:union}, it consistently outperforms other models on MeteoNet and HKO-7.

\begin{figure}[!ht]
    \centering
    \includegraphics[width=1\linewidth]{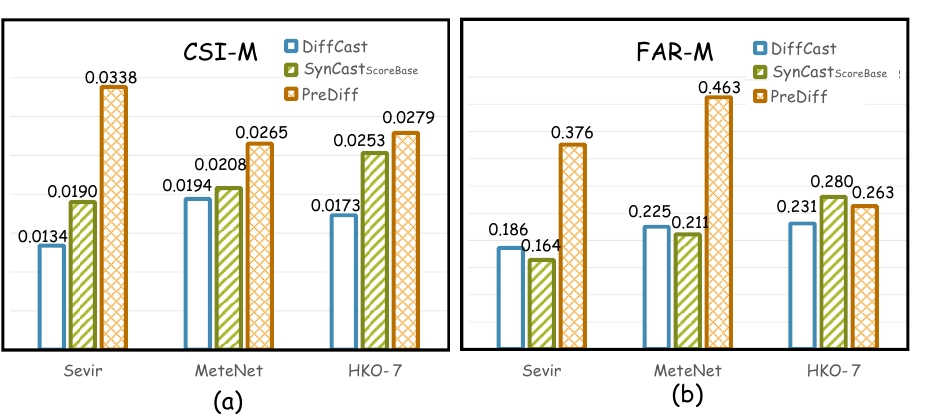}
    \caption{Comparison of the average differences in CSI-M and FAR-M between preference pairs across three radar precipitation datasets for DiffCast, $\text{SynCast}_{\text{ScoreBase}}$, and PreDiff.}
    \label{fig:Diff}
\end{figure}

\subsection{Generalization and Effectiveness of Diffusion-SPO}
\subsubsection{Diffusion-SPO on Score-Based Diffusion Models}
To verify the effectiveness and generalization of Diffusion-SPO, we apply it to three kinds of score-based diffusion models designed for precipitation nowcasting. 
These include the deterministic-hybrid model DiffCast, DDPM-sampled model PreDiff, and DDIM-sampled model $\text{SynCast}_{\text{ScoreBase}}$ (i.e., the SynCast model without Diffusion-SPO), which become SynDiffCast, SynPreDiff, and SynCast after Diffusion-SPO fine-tuning, respectively.
As shown in Table~\ref{tab:cross_model}, these models generally improved both CSI and FAR in most cases after Diffusion-SPO optimization. 
SynPreDiff showed the most significant improvement, achieving a 1.87\% decrease in FAR on HKO-7, along with a 10.75\% and 10.74\% increase in CSI-M-POOL4 and CSI-M-POOL16, respectively, followed by SynCast and SynDiffCast.
Such improvements are closely related to model's stochasticity.
Fig.~\ref{fig:Diff} presents the average differences in CSI-M and FAR-M between win and lose samples for the three models across the datasets. The observed ranking: $ \text{PreDiff} > \text{SynCast}_{\text{ScoreBase}} > \text{DiffCast}$ aligns with the magnitude of CSI improvements.

Intuitively, models with higher stochasticity could generate more diverse samples, which increases the gap between win and lose samples and provides clearer preference signals for Diffusion-SPO. 
The above three models differ in stochasticity. 
Specifically, DDPM involves more sampling steps with random noise added at each step, resulting in the highest stochasticity. 
DDIM, in contrast, uses fewer steps and is typically deterministic, exhibiting lower stochasticity. 
DiffCast, which is a deterministic model conditioned on its inputs, has the lowest stochasticity of the three.
It further highlights the decisive role of model stochasticity in Diffusion-SPO optimization.

\subsubsection{Diffusion-SPO on Flow-Matching Diffusion Models}
Furthermore, we applied Diffusion-SPO to $\text{SynCast}_{\text{FlowBase}}$, which was trained using Flow Matching~\cite{lipman2022flow}. As shown in Table~\ref{tab:cross_model2}, $\text{SynCast}_{\text{Flow}}$ also exhibited improvements in both CSI and FAR, demonstrating the effectiveness of Diffusion-SPO across different diffusion-based generative mechanisms. However, Flow Matching represents an almost deterministic mapping, so its stochasticity is lower than that of score-based methods. As a result, its performance gains are slightly smaller compared to DDPM or DDIM sampled models.

\subsubsection{Visual Interpretation of Diffusion-SPO}
\begin{figure*}[ht]
    \centering
    \includegraphics[width=0.87\linewidth]{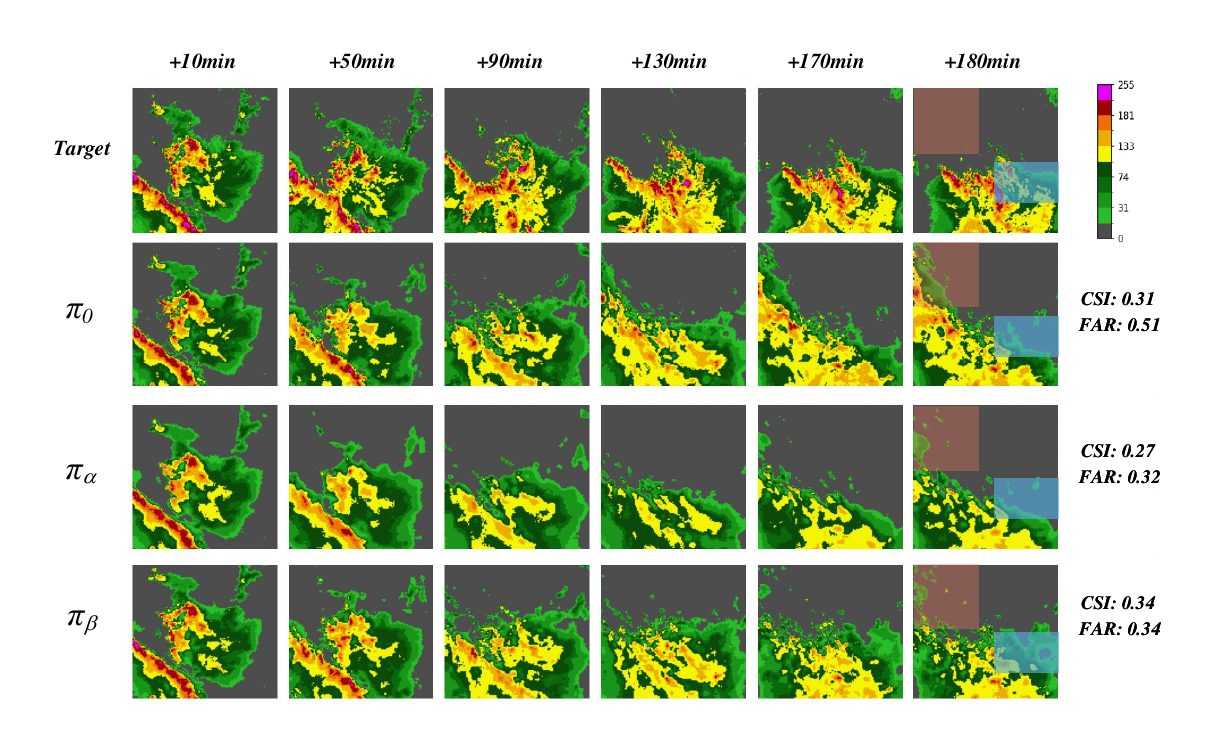}
    \caption{Qualitative visualization of SynCast's FAR and CSI preference optimization in Diffusion-SPO.}
    \label{fig:3stage}
\end{figure*}

To better illustrate the effect of Diffusion-SPO, we visualize the generated results of the three models $\pi_0$, $\pi_{\alpha}$, and $\pi_{\beta}$, as shown in Fig.~\ref{fig:3stage}. 
We observe that during 0 - 90 min, $\pi_0$ largely preserves the overall precipitation pattern. But in the later stage, a clear false alarm appears in the red region at the upper-left corner. 
With FAR preference optimization, $\pi_{\alpha}$ effectively suppresses this false alarm, but a miss still persists in the blue region at the lower-right corner.
Finally, incorporating CSI preference optimization enables $\pi_{\beta}$ to recover previously missed precipitation and the constraint aligning with FAR in the second stage prevents false alarms from reappearing.

\begin{table}[t]
\centering
\caption{Diffusion-SPO Post-training Results on SEVIR for Score-based and Flow Matching Diffusion Models. }
\label{tab:cross_model2}
\begin{tabular}{ccccc}
\toprule
\multirow{2}{*}{Model} & \multirow{2}{*}{FAR-M} & \multicolumn{3}{c}{CSI-M} \\
\cmidrule(lr){3-5}
 & & POOL1 & POOL4 & POOL16 \\
\midrule
$\text{SynCast}_{\text{FlowBase}}$ & 0.606 & 0.238 & 0.271 & 0.301 \\
\multirow{2}{*}{$\text{SynCast}_{\text{Flow}}$} 
  & 0.587 & 0.245 & 0.278 & 0.303 \\ 
  & \textbf{(-3.13\%)} &\textbf{ (+2.94\%)} & \textbf{(+2.58\%)} & \textbf{(+0.66\%) } \\
\midrule
$\text{SynCast}_{\text{ScoreBase}}$ & 0.526 & 0.242 & 0.269 & 0.294 \\
\multirow{2}{*}{$\text{SynCast}_{\text{Score}}$}  
  & 0.505 & 0.251 & 0.282 & 0.305 \\  
  & \textbf{(-3.99\%)} & \textbf{(+3.71\%)} & \textbf{(+4.83\%)} & \textbf{(+3.74\%)} \\
\bottomrule
\end{tabular}
\end{table}

\begin{table*}[t]
\centering
\caption{Comparison of different preference optimization (PO) methods on SEVIR, MeteoNet, and HKO-7 datasets.}
\label{tab:po_ablation}
\begin{tabular}{lcccccccccccc}
\toprule
\multirow{2}{*}{PO Ways} & \multicolumn{4}{c}{SEVIR} & \multicolumn{4}{c}{MeteoNet} & \multicolumn{4}{c}{HKO-7} \\
\cmidrule(lr){2-5} \cmidrule(lr){6-9} \cmidrule(lr){10-13}
 & FAR-M & POOL1 & POOL4 & POOL16 & FAR-M & POOL1 & POOL4 & POOL16 & FAR-M & POOL1 & POOL4 & POOL16 \\
\midrule
Diffusion-DPO & 0.511 & 0.251 & 0.281 & 0.305 & 0.463 & 0.240 & 0.279 & 0.281 & 0.491 & 0.245 & 0.306 & 0.357 \\
Diffusion-SPO & \textbf{0.505} & \textbf{0.251} & \textbf{0.282} & \textbf{0.305} &\textbf{ 0.461} &\textbf{ 0.242} & \textbf{0.282} & \textbf{0.284} & \textbf{0.491} & \textbf{0.251} & \textbf{0.313} & \textbf{0.363} \\
\bottomrule
\end{tabular}
\end{table*}

\begin{table*}
\centering
\caption{Performance Comparison of Sample Selection Strategies (Dual-Metric, Whole-sample, and Frame-level) on SEVIR, MeteoNet, and HKO-7 Datasets}
\label{tab:sample_ablation}
\begin{tabular}{lcccccccccccc}
\toprule
\multirow{2}{*}{Sample Ways} & \multicolumn{4}{c}{SEVIR} & \multicolumn{4}{c}{MeteoNet} & \multicolumn{4}{c}{HKO-7} \\
\cmidrule(lr){2-5} \cmidrule(lr){6-9} \cmidrule(lr){10-13}
 & FAR-M & POOL1 & POOL4 & POOL16 & FAR-M & POOL1 & POOL4 & POOL16 & FAR-M & POOL1 & POOL4 & POOL16 \\
\midrule
Dual-Metric & 0.517 & 0.249 & 0.277 & 0.298 & 0.465 & 0.240 & 0.279 & 0.282 & 0.499 & 0.245 & 0.301 & 0.342 \\
Whole-sample & 0.510 & 0.249 & 0.279 & 0.301 & 0.464 & 0.238 & 0.278 & 0.273 & 0.495 & 0.248 & 0.310 & 0.356 \\
Frame-level & \textbf{0.505} & \textbf{0.251} & \textbf{0.282} & \textbf{0.305} &\textbf{ 0.461} &\textbf{ 0.242} & \textbf{0.282} & \textbf{0.284} & \textbf{0.491} & \textbf{0.251} & \textbf{0.313} & \textbf{0.363} \\
\bottomrule
\end{tabular}
\end{table*}

\subsection{Analysis and Ablation Study}
\subsubsection{Preference Optimization Methods}
In this section, we conduct ablation experiments on different preference optimization strategies. 
Specifically, we compare two methods: (i) a straightforward two-stage Diffusion-DPO without any additional constraints; and (ii) our proposed Diffusion-SPO method. 
Results in Table~\ref{tab:po_ablation} show that both methods achieve comparable CSI-M performance on the SEVIR dataset, while Diffusion-SPO attains a relatively lower FAR.
This indicates that despite the inherent conflict between CSI and FAR, Diffusion-SPO better maintains the lower FAR achieved in the first stage while optimizing for CSI.
Overall, these findings demonstrate the advantage of Diffusion-SPO in achieving simultaneous improvements across conflicting metrics.

\subsubsection{Preference Pairs Construction}
We compare three strategies for constructing win/lose preference pairs:
\begin{itemize}
    \item \textbf{Dual-metric}: We rank candidate sequences by both CSI and FAR, selecting those that rank among the top two for both metrics as win samples, and those that rank among the bottom two as lose samples. However, due to the inherent conflict between CSI and FAR, such sequences are relatively scarce;
    \item \textbf{Whole-sample}: We directly compare the overall CSI and FAR scores of the entire sequence, selecting the best and worst sequences;
    \item \textbf{Frame-level}: For each candidate sequence, we identify the best and worst frames with respect to CSI and FAR, and then assemble them into complete sequence samples.

\end{itemize}

Results are reported in Table~\ref{tab:sample_ablation}. We observe that the Frame-level strategy performs best, which can be attributed to three factors:
(i) Finer-grained supervision allows the model to capture frame-level variations that obscured by the coarse comparisons in the whole-sample strategy;
(ii) It overcomes the dual-metric strategy's limitation of limited valid samples due to stringent constraints;
(iii) Frame-level processing promotes higher diversity, leading to more distinctive and informative preference pairs.

\section{Conclusion and Future Work}
Despite advances in probabilistic models for precipitation nowcasting, challenges remain in consistently reaching the upper bounds of performance and   
balancing conflicting metrics. To address this, we introduce reinforcement learning for preference-aligned optimization in precipitation nowcasting for the first time.
Specifically, we propose SynCast, a precipitation nowcasting model using a post-training Diffusion-SPO framework to balance conflicting evaluation metrics.
SynCast first trains a highly stochastic diffusion model and generates multiple prediction sequences for each input by varying the initial noise. For each frame, CSI and FAR are computed, and frames with the best and worst performances are combined to form ``win” and ``lose” samples. 
In the post-training stage, a two-stage Diffusion-SPO is adopted. We first perform alignment on the FAR dimension using Diffusion-DPO, and then optimize CSI under the FAR-alignment constraint. 
Future work will focus on extending SynCast to higher-resolution predictions and more conflicted metrics.

\bibliographystyle{IEEEtran}
\bibliography{ref}










\vspace{11pt}
\begin{IEEEbiography}
[{\includegraphics[width=1in,height=1.25in,clip,keepaspectratio]{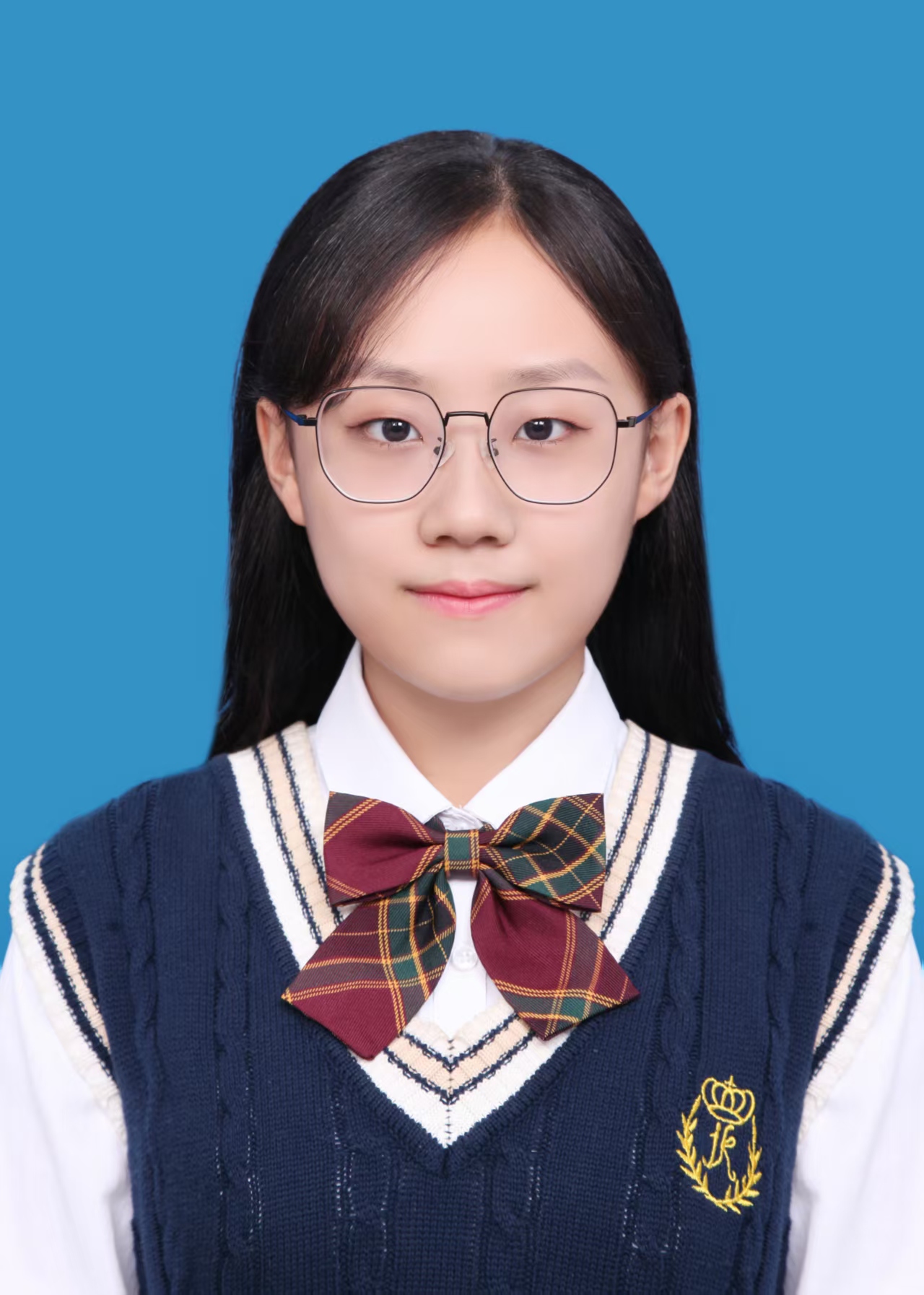}}] 
{Kaiyi Xu} is currently pursuing the Ph.D. degree with the School of Information Science and Technology, University of Science and Technology of China (USTC), Anhui, China. She received the B.S. degree in Computer Science from the China University of Geosciences, Wuhan, China. Her research interests include AI for Science and generative models.
\end{IEEEbiography}

\begin{IEEEbiography}
[{\includegraphics[width=1in,height=1.25in,clip,keepaspectratio]{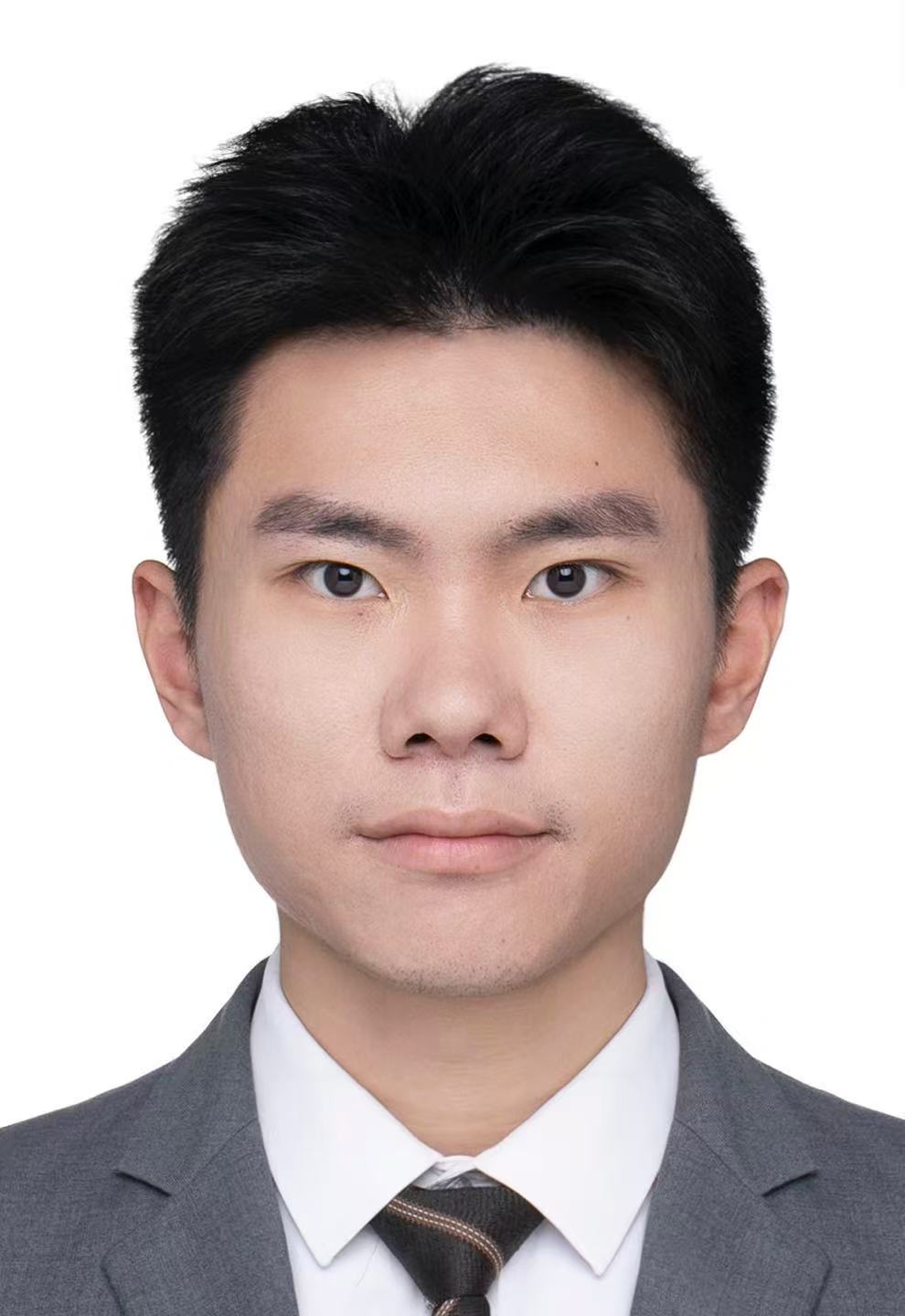}}] 
{Junchao Gong} is a fourth-year Ph.D. student in the Department of Information and Communication Engineering at Shanghai Jiao Tong University. He received his B.S. degree from the School of Computer Science at Shanghai Jiao Tong University in 2022 and is currently pursuing his doctoral studies there. He has published several peer-reviewed papers in top-tier venues, including ICML, ICLR and NeurIPS. His research interests span spatiotemporal learning, weather prediction, and radar extrapolation.
\end{IEEEbiography}

\begin{IEEEbiography}
[{\includegraphics[width=1in,height=1.25in,clip,keepaspectratio]{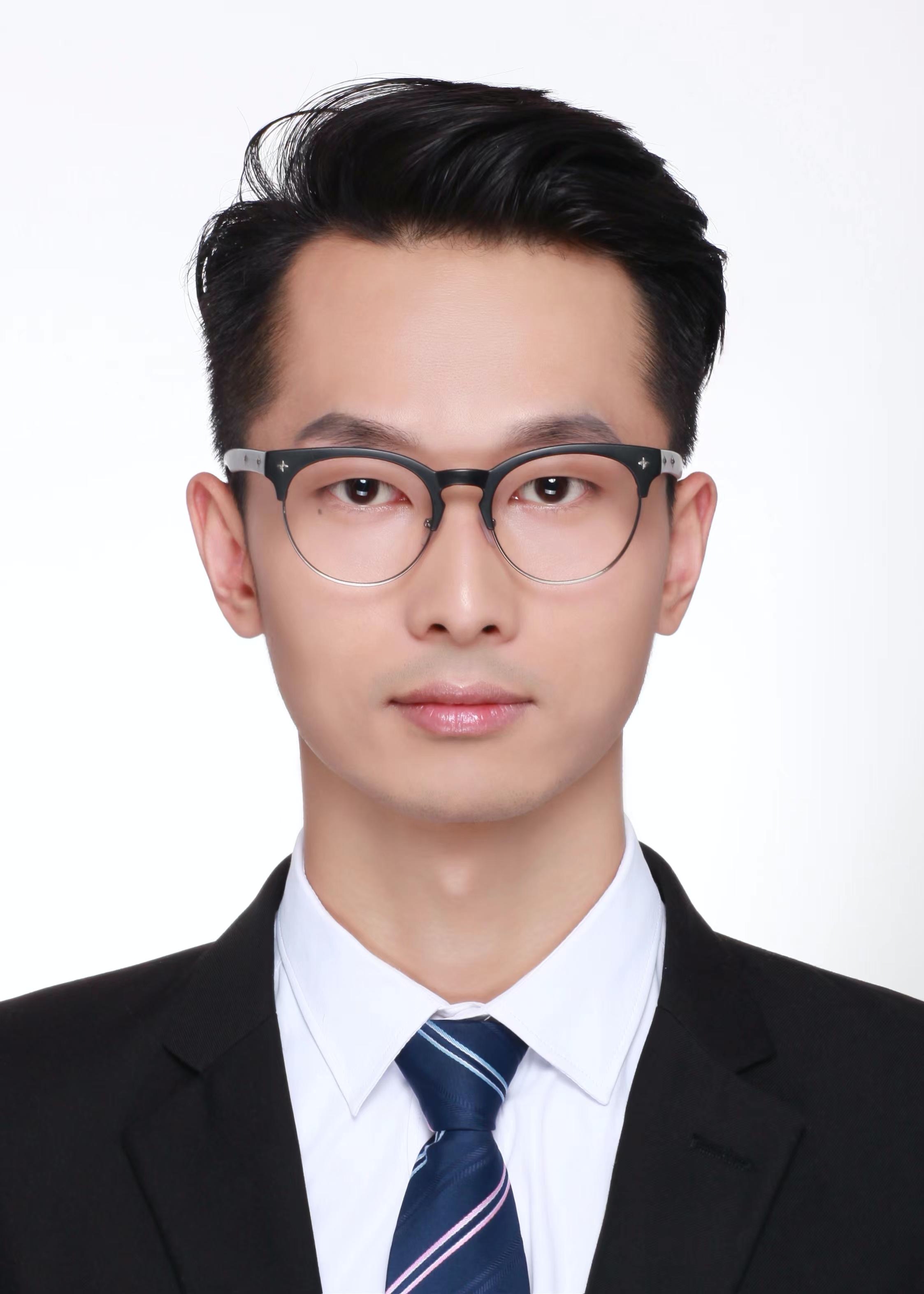}}] 
{Wenlong Zhang} received the B.E. degree from the Zhengzhou University of Light Industry, Zhengzhou, China, in 2016, and the M.S. degree from the Beijing Institute of Technology, Beijing, China, in 2018. He received the Ph.D. degree from The Hong Kong Polytechnic University, Hong Kong, China, in 2024. He is currently a young researcher of Shanghai AI Laboratory, Shanghai, China. His
research interests include machine learning and AI for Science.
\end{IEEEbiography}



\begin{IEEEbiography}
[{\includegraphics[width=1in,height=1.25in,clip,keepaspectratio]{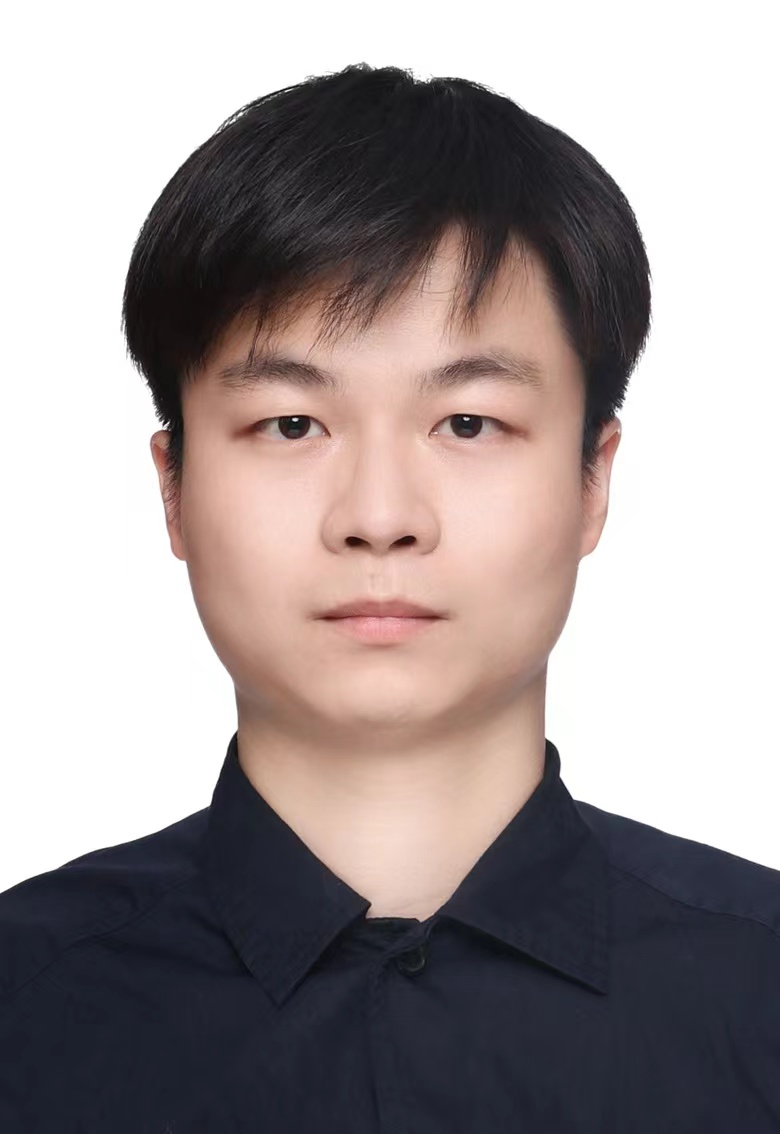}}] 
{Ben Fei} received the M.S. degree in the Department of Materials Science from Fudan University, Shanghai, China, in 2021. 
He received a Ph.D. degree in the School of Computer Science at Fudan University.
He is currently a postdoctoral fellow at Multimedia Lab, Department of Information Engineering, The Chinese University of Hong Kong, Hong Kong SAR.
His research interests include generative models, 3D computer vision and AI for Science. 
He is an IEEE Young Professional.
\end{IEEEbiography}

\begin{IEEEbiography}
[{\includegraphics[width=1in,height=1.25in,clip,keepaspectratio]{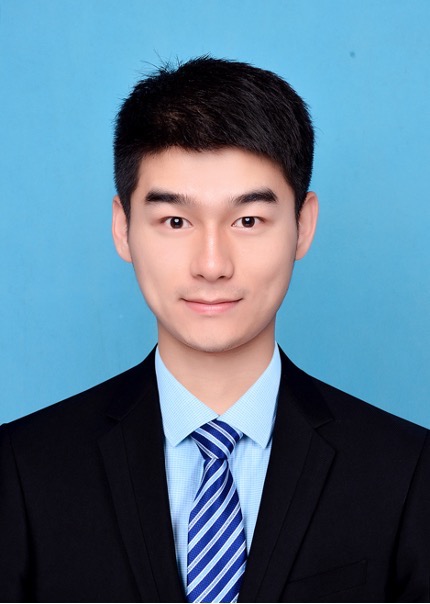}}] 
{Lei Bai (Member, IEEE)} received the PhD degree from UNSW, Sydney, in 2021. He is currently a Research Scientist at Shanghai Artificial Intelligence Laboratory. His research interests include Spatiotemporal Learning, Computer Vision, Multi-Agents System, and their applications in inter-disciplinary research (e.g., earth science, urban science). Lei has authored or coauthored more than 80 papers in top-tier AI conferences and journals, including Nature Machine Intelligence, Nature Communications, IEEE Transactions on Pattern Analysis and Machine Intelligence, NeurIPS, ICML, and CVPR. He also regularly serves as a program committee member or reviewer for these journals and conferences. Lei served as an Area Chair for PRCV 2023/2024 and a Workshop Chair for DICTA 2022 and VALSE 2023. He was the recipient of the 2024 IEEE TCSVT Best Paper Award, 2022 WAIC Yunfan Award, and 2019 Google PhD Fellowship.
\end{IEEEbiography}

\begin{IEEEbiography}
[{\includegraphics[width=1in,height=1.25in,clip,keepaspectratio]{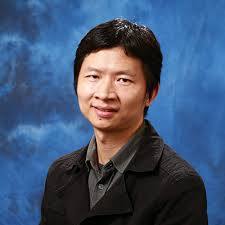}}] 
{Wanli Ouyang (Senior Member, IEEE)} received
the Ph.D. degree from the Department of Electronic Engineering, The Chinese University of
Hong Kong (CUHK). He is currently an Associate
Professor with the School of Electrical and Information Engineering, The University of Sydney,
Sydney, Australia. His research interests include
image processing, computer vision, and pattern
recognition.
\end{IEEEbiography}



\end{document}